\documentclass{article}

\usepackage{microtype}
\usepackage{graphicx}
\usepackage{subfig}
\usepackage{booktabs}

\usepackage{hyperref}

\usepackage[accepted]{icml2024} 

\usepackage{amsmath}
\usepackage{amssymb}
\usepackage{mathtools}
\usepackage{amsthm}

\usepackage[capitalize,noabbrev]{cleveref}

\theoremstyle{plain}

\theoremstyle{definition}

\theoremstyle{remark}

\usepackage[textsize=tiny]{todonotes}

\usepackage{xcolor}

\icmltitlerunning{Self-Improved Learning for Scalable Neural Combinatorial Optimization}

\begin{document}

\twocolumn[
\icmltitle{Self-Improved Learning for Scalable Neural Combinatorial Optimization}


\begin{icmlauthorlist}
\icmlauthor{Fu Luo}{Sustech}
\icmlauthor{Xi Lin}{CityU}
\icmlauthor{Zhenkun Wang}{Sustech}
\icmlauthor{Xialiang Tong}{Huawei}
\icmlauthor{Mingxuan Yuan}{Huawei}
\icmlauthor{Qingfu Zhang}{CityU}

\end{icmlauthorlist}

\icmlaffiliation{Huawei}{Huawei Noah’s Ark Lab, Shenzhen, China.}
\icmlaffiliation{Sustech}{Southern University of Science and Technology, Shenzhen, China.}
\icmlaffiliation{CityU}{City University of Hong Kong, Hong Kong SAR.}

\icmlcorrespondingauthor{Zhenkun Wang}{ wangzhenkun90@gmail.com}


\vskip 0.3in
]

\printAffiliationsAndNotice{} 

\begin{abstract}
The end-to-end neural combinatorial optimization (NCO) method shows promising performance in solving complex combinatorial optimization problems without the need for expert design. However, existing methods struggle with large-scale problems, hindering their practical applicability. To overcome this limitation, this work proposes a novel Self-Improved Learning (SIL) method for better scalability of neural combinatorial optimization. Specifically, we develop an efficient self-improved mechanism that enables direct model training on large-scale problem instances without any labeled data. Powered by an innovative local reconstruction approach, this method can iteratively generate better solutions by itself as pseudo-labels to guide efficient model training. In addition, we design a linear complexity attention mechanism for the model to efficiently handle large-scale combinatorial problem instances with low computation overhead. Comprehensive experiments on the Travelling Salesman Problem (TSP) and the Capacitated Vehicle Routing Problem (CVRP) with up to 100K nodes in both uniform and real-world distributions demonstrate the superior scalability of our method.
\end{abstract}

\begin{figure}[t]
\centering
\includegraphics[width=1\columnwidth]{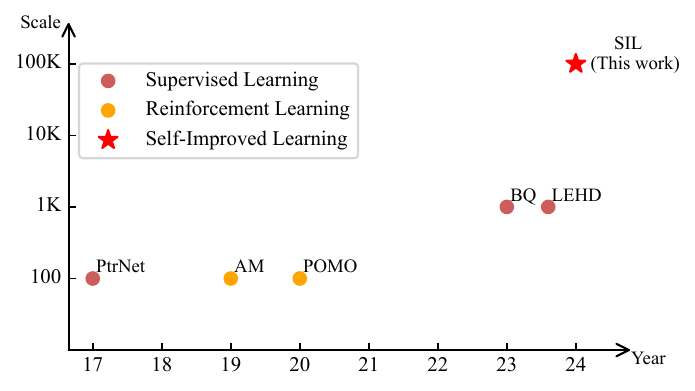}
\caption{Development of constructive NCO methods.}
\label{fig_construction_nco_problem_size}
\end{figure}

\section{Introduction}
\label{Introduction}
Combinatorial optimization (CO) is crucial in numerous real-world applications, such as transportation~\citep{garaix2010vehicle}, navigation~\citep{elgarej2021optimized}, and circuit design~\citep{brophy2014principles}. Due to the NP-hard nature, many CO problems are challenging to solve~\citep{ausiello2012complexity}. While different methods have been developed to tackle various CO problems over the past decades, they typically suffer from two major limitations: the requirement for profound domain expertise leads to substantial development costs, and the high computational complexity hampers their practicality on large-scale problems.

The recent development of machine learning has led to the emergence of the neural combinatorial optimization (NCO) method for solving complex CO problems in an end-to-end manner~\citep{bengio2021machine}. This method builds a powerful model to autonomously learn problem-solving strategies from data, eliminating the need for costly manual algorithm design. Moreover, the learned model can directly generate an approximate solution for a given CO problem instance with a low computational cost. In this way, NCO methods exhibit great potential to overcome the two major limitations of traditional approaches for solving CO problems. However, despite their promising performance in solving small-scale problems, existing NCO methods often struggle when applied to problems with more than 1K nodes as shown in Figure~\ref{fig_construction_nco_problem_size}, thereby restricting their practical applicability.

\begin{figure*}[t]
\centering
\subfloat[Supervised Learning]{\includegraphics[width = 0.33\linewidth]
{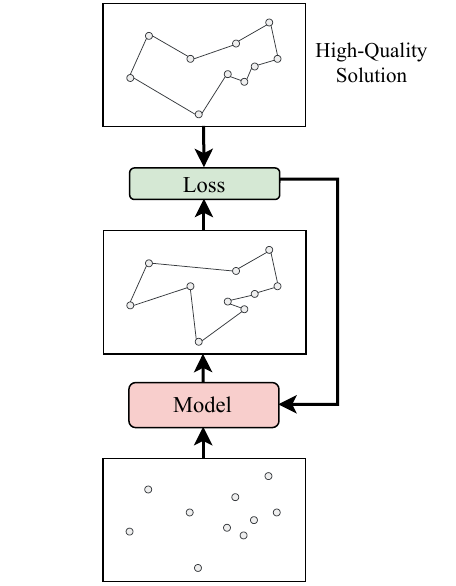}} \hspace{0mm}
\subfloat[Reinforcement Learning] {\includegraphics[width = 0.30\linewidth]{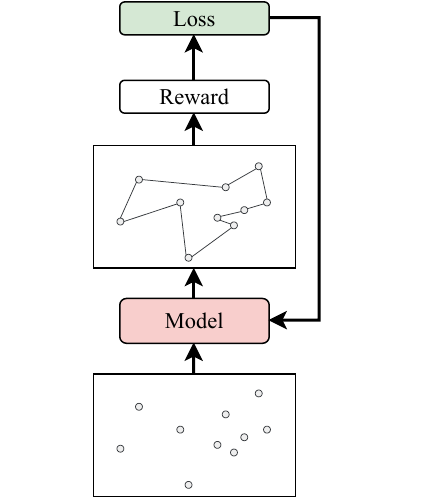}}\hspace{0mm}
\subfloat[Self-Improved Learning (This Work)]{\includegraphics[width = 0.36\linewidth]{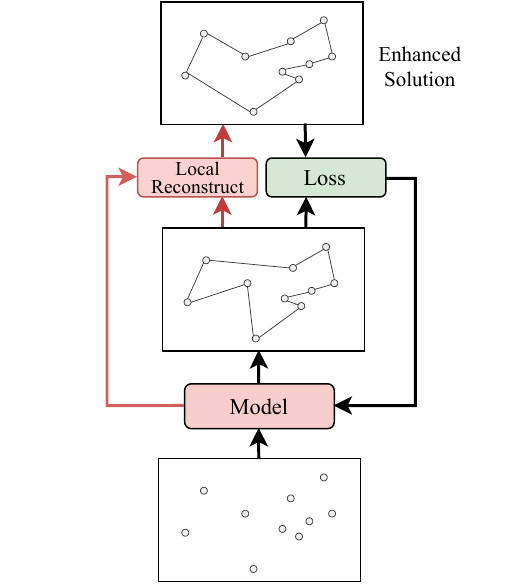}}
\caption{\textbf{Different Learning Methods for NCO.} \textbf{(a) Supervised Learning (SL) :} SL-based method relies on high-quality solutions for model training. Its applicability to large-scale problems is restricted by the difficulty of obtaining high-quality solutions. \textbf{(b) Reinforcement Learning (RL):} RL-based method requires the generation of complete solutions to calculate rewards during model training. For large-scale problems, it is hindered by the issues of sparse rewards and high computational costs. \textbf{(c) Self-Improved Learning (SIL, This Work)}: SIL has a novel iterative cycle that contains 1) a local reconstruction step to produce enhanced solutions for model training, and 2) a model training step to further strengthen the local reconstruction performance. In this way, SIL can tackle large-scale problem with up to 100K nodes.
}

\label{fig_learning_method_comparison}
\end{figure*}

Directly learning an NCO model on the original large-scale problem is very tough, of which the fundamental challenge comes from the current training approaches, namely supervised learning (SL) and reinforcement learning (RL). SL-based approaches are restricted by the difficulty of obtaining sufficient labels (i.e., high-quality solutions) for large-scale problems, while RL-based approaches struggle with the extremely sparse reward. Moreover, these methods typically learn to construct the complete solution for a large-scale problem instance, which entails processing reams of information simultaneously, resulting in substantial computational costs and overwhelming the model's learning capabilities. In addition to direct training, some methods attempt to train NCO models on small-scale problems and then generalize them to solve problems with much larger sizes. However, even the current state-of-the-art models such as BQ~\citep{drakulic2023bq} and LEHD~\citep{luo2023neural} are unable to effectively handle large-scale CO problem instances with 10K-100K nodes. Another attempt is to decompose a large-scale problem into multiple smaller subproblems, of which both the problem decomposition and subproblem solving steps are conducted by models~\cite{hou2023generalize,ye2024glop}. This method transfers the difficulty to the problem decomposition that is also NP-hard, and cannot essentially overcome the challenge~\citep{edelkamp2011heuristic}. Problem decomposition may even change the property of the original problem, causing an unconquerable optimality gap. Furthermore, it destroys the correlation between the subproblems, which is not conducive to taking full advantage of deep learning models.

This work proposes a novel Self-Improved Learning (SIL) method for scalable NCO to solve large-scale CO problems with up to 100K nodes. Our contributions can be summarized as follows:
\begin{itemize}
    \item We develop a novel and efficient self-improved learning method that allows NCO models to be directly trained on large-scale CO problem instances without any labeled data. 
    \item We design a linear complexity attention mechanism for NCO models to handle large-scale problems, which significantly reduces the computational cost for both model training and inference. 
    \item We conduct comprehensive experiments on both randomly generated and real-world Travelling Salesman Problem (TSP) and Capacitated Vehicle Routing Problem (CVRP) benchmarks. The results fully demonstrate that our proposed SIL method can achieve state-of-the-art performance on large-scale VRPs with up to 100K nodes. 
\end{itemize}

\section{Related Works}

\paragraph{Constructive NCO.}
The constructive NCO method, also known as the end-to-end method, learns a model to construct the approximate solutions for given problem instances in an autoregressive manner. Pioneering works~\citep{vinyals2015pointer,bello2016neural,nazari2018reinforcement} show that RNN-based neural network models, trained with supervised learning (SL) or reinforcement learning (RL), can achieve promising results on small-scale CO problems. \citet{kool2018attention} and \citet{deudon2018learning} leverage the Transformer structure~\cite{vaswani2017attention} to develop powerful attention-based models to solve a diverse set of small-scale routing problems, which mark significant advancements in constructive NCO. Since then, various Transformer-based methods have been proposed with different improvements~\citep{xin2021multi,xin2020step,kwon2020pomo,hottung2021efficient,kim2021learning,choo2022simulation,kim2022sym}. However, they typically only perform well on small-scale problems with up to 100 nodes. Some works attempt to train the NCO model to tackle large-scale problems by iteratively constructing partial solutions~\citep{kim2021learning,pan2023h-tsp,cheng2023select,ye2024glop}. Nevertheless, due to the limitation of RL-based learning, they only perform training on partial solutions of small-scale and fixed-size, which impedes their scalability on large-scale problems. Current developments such as BQ~\citep{drakulic2023bq} and LEHD~\citep{luo2023neural}, trained with SL, have extended their applicability to CO problems with up to 1K nodes. However, due to the lack of high-quality labeled data on large-scale problems, these SL-based methods perform poorly on problems with more than 10K nodes. 

\paragraph{Two-stage NCO.} 
This method first builds a graph neural network (GNN) model to predict a heatmap that measures the probability for each edge to be in the optimal solution, and then iteratively searches for an approximate solution using the heatmap~\citep{joshi2019efficient}. Different methods such as beam search~\citep{joshi2019efficient}, dynamic programming~\citep{kool2022deep} and Monte Carlo Tree Search (MCTS)~\citep{fu2021generalize} have been used for solution searching. Some recent works are proposed to address large-scale TSP instances~\citep{fu2021generalize,qiu2022dimes,sun2023difusco}. Since they rely on the search strategy specifically designed for TSP, they cannot be applied to solve other complicated CO problems such as CVRP. 

\paragraph{Decomposition-based Methods.}
These methods decompose a large-scale problem into multiple simpler small-scale subproblems to solve, and then merge the solutions of all small-scale subproblems to construct the complete solution for the original large-scale problem~\citep{li2021learning,zong2022rbg,hou2023generalize}. Although they perform well in solving large-scale TSP instances, they cannot be used to solve other complicated CO problems, such as CVRP. The decomposition of these problems becomes fiendishly intractable and cannot be achieved via a universal model or strategy. Moreover, the problem decomposition overlooks the correlation between subproblems and may change the optimality of the problem, leading to suboptimal solutions.

\begin{figure*}[t]
\centering
\includegraphics[width=1\textwidth]{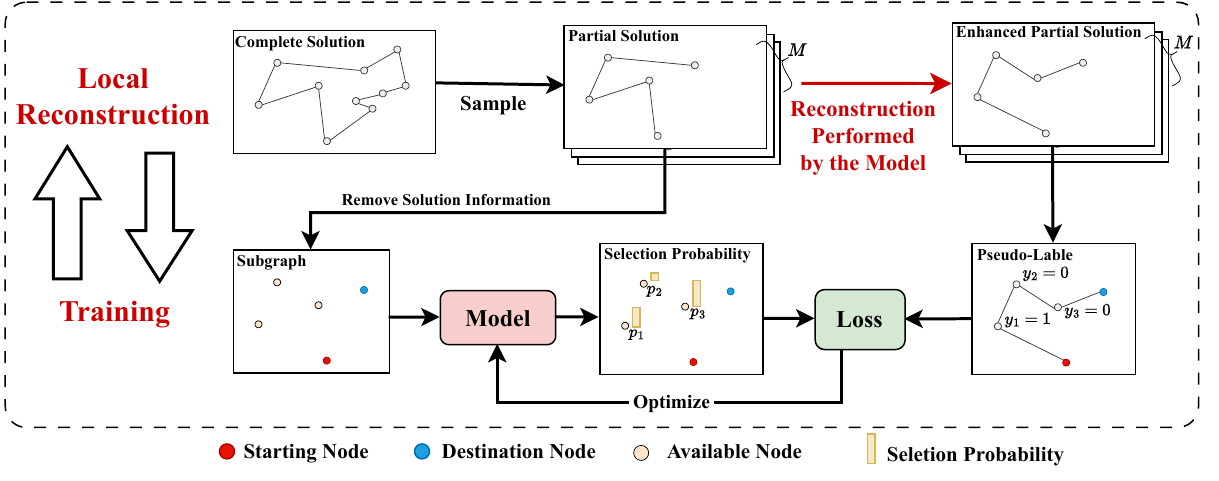}
\caption{The self-improved learning (SIL) process. It involves a cycle of iterative self-improvement. In each interaction, the model performs a local reconstruction to improve the solution quality. Then the enhanced solutions act as pseudo-labels for the model training to improve model performance.}
\label{Overview of the training method}
\end{figure*}

\section{Preliminaries}

This work focuses on the constructive NCO method, which typically employs an encoder-decoder model for solution construction. The encoder processes node features (e.g., coordinates) to produce node embeddings. Then, based on the node embeddings, the decoder sequentially constructs the solution by adding one node at each step. Initially, all nodes are available for the decoder to select. However, as construction proceeds, the selected nodes become unavailable. At each step, the decoder predicts the probability of selecting each available node and chooses one based on these probabilities. The selected node is then appended to the current partial solution. The process continues until no node is available for selection, and the resulting complete solution is returned as the model's output.

\section{Self-Improved Learning}
In this work, we propose a novel self-improved learning (SIL) method to train NCO models on large-scale problems. As illustrated in Figure~\ref{Overview of the training method}, SIL involves an iterative cycle of local reconstruction approach and model training. The local reconstruction step produces enhanced solutions to guide model training, while the improved model further strengthens the local reconstruction approach to generate even better solutions. In this way, SIL can continuously boost the performance of the NCO model in solving large-scale problems with up to 100K nodes.  
\subsection{Local Reconstruction}
Constructive NCO models exhibit bias in decoding, where variations in starting nodes, destination nodes, and directions can result in vastly different solutions~\citep{kwon2020pomo}. Benefits from this bias, the model can gradually improve the solution quality by performing iterative local reconstruction during the inference phase~\citep{luo2023neural,ye2024glop}. Our proposed SIL innovatively leverages this local reconstruction for NCO model training, significantly boosting the model scalability from problems with 1K nodes to those with 100K nodes.

Our proposed parallel local reconstruction process is illustrated in Figure \ref{Overview of the training method} with a single solution as an example. Given a $n$-node TSP instance, we denote its solution as $\boldsymbol{\tau} = (x_1, x_2, \ldots, x_n)^\intercal$. A partial solution $\boldsymbol{\tau}^{p}$ is defined as a contiguous subset of $\boldsymbol{\tau}$. In the first step, $M$ non-overlapping partial solutions of size $\omega$ are uniformly sampled from $\boldsymbol{\tau}$. As $\boldsymbol{\tau}$ can be expressed as a circle, we randomly choose the sampling direction, clockwise or counterclockwise. The length $\omega$ is randomly determined from the range $[4, l_{max}]$, where $l_{max}$ is a parameter used to decide the maximum size of the partial solution. In this paper, $M$ is set to $\lfloor l_{max}/\omega \rfloor$. 

In the second step, the model reconstructs each partial solution $\boldsymbol{\tau}^p_i$ node by node from its starting node to its destination node, i.e., rearranges the order of the nodes within the starting node and the destination node for each partial solution. The yielded new partial solution $\boldsymbol{\tau}^{p\prime}_{i}$ is compared with $\boldsymbol{\tau}^p_i$, and the better one (e.g., the one with shorter length) is retained. The retained $M$ partial solutions are merged to $\boldsymbol{\tau}^{p}$ to form a complete solution. Through iterative local reconstructions, the solution quality can be continuously improved. The improved solutions can in turn serve as pseudo-labels to guide the model training more efficiently.

\subsection{Training to Improve Model Performance}

Our model training strategy is similar to LEHD's that learns to construct partial solutions~\citep{luo2023neural}. 
The training process of SIL is also illustrated in Figure~\ref{Overview of the training method}. For an enhanced solution gained via the local reconstruction $\boldsymbol{\tau}$, we sample a partial solution $\boldsymbol{\tau}^p$ from it as pseudo-labels and let the model predict the order for the nodes involved in $\boldsymbol{\tau}^p$ in an autoregressive manner. The cross-entropy loss is used to measure the prediction error, i.e.,
\begin{equation}
    \mathcal{L} = -\sum_{i=1}^{u} y_i \log(p_i),
\end{equation}
where $p_i$ denotes the selection probability of node $\mathbf{s}_i$ and $u$ represents the count of available nodes. 
The training process terminates if the enhanced solutions, serving as pseudo-labels, cannot be used to improve model performance further or a predefined budget is reached. Subsequently, the local reconstruction is conducted using the trained model to generate new enhanced solutions. Thereafter, a new cycle of self-improved learning is operated.

\begin{figure*}[t]
\centering
\subfloat[Idea of the proposed linear attention mechanism.]{\includegraphics[width = 0.49\linewidth]
{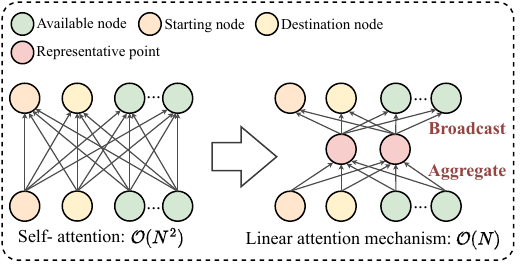}} \hspace{2mm}
\subfloat[$l$-th linear attention module.] {\includegraphics[width = 0.27\linewidth]{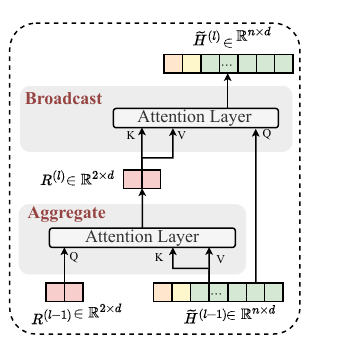}}\hspace{2mm}
\subfloat[The model structure.]{\includegraphics[width = 0.20\linewidth]{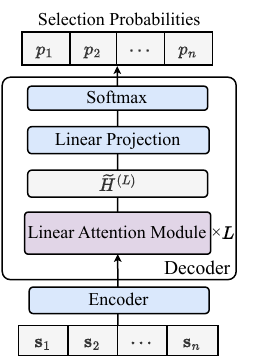}}
\caption{Linear attention model design. The proposed linear attention mechanism utilizes a certain number of representative points to aggregate key information about the graph and broadcast this information to all nodes, which efficiently eliminates the need for explicit computations between all nodes during the attention process, thereby achieving linear complexity.}

\label{Model Structure Overview}
\end{figure*}
\section{Linear Attention Model}
It is infeasible to directly apply self-improved learning to existing NCO models on large-scale problems, as they employ a quadratic attention mechanism with high complexity. This work proposes a linear attention mechanism to overcome this challenge. 

\subsection{Linear Attention Mechanism}
The self-attention mechanism is frequently used in the decoder to dynamically relate the starting, destination, and available nodes, as well as assign a selection probability to each available node during the whole decoding process. However, as illustrated in Figure~\ref{Model Structure Overview}(a), the self-attention operator incurs a high computational complexity of $\mathcal{O}(N^2)$, which is caused by the exhaustive pairwise interactions between nodes.    

This work designs a lightweight linear attention mechanism to integrate information among nodes with much lower computational complexity. As illustrated in Figure \ref{Model Structure Overview}(a), this approach utilizes a certain number of representative points to \textbf{aggregate} key information about the graph, such as the starting and destination nodes of the current route, nearby nodes of the starting node, and so on. These representative points then \textbf{broadcast} this information to all nodes, which enables more efficient information updates. 

\subsection{Model Structure}
Our proposed model also uses an encoder-decoder structure like other NCO models. Notably, its decoder features $L$ stacked linear attention modules as shown in Figure \ref{Model Structure Overview}(c), and the detailed structure of the linear attention module is illustrated in Figure \ref{Model Structure Overview}(b).  
\paragraph{Encoder.} For a problem instance $\mathbf{s}$ with node features $(\mathbf{s}_1,\ldots,\mathbf{s}_n)$ (e.g., coordinates of $n$ cities), the encoder transforms each node feature $\mathbf{s}_i$ into node embedding $\mathbf{h}_i\in \mathbb{R}^{d}$ using a linear projection, i.e.,
\begin{equation}
    \mathbf{h}_i={W}^{(0)}\mathbf{s}_i+\mathbf{b}^{(0)}, \text{for} i =1,\ldots,n,
\end{equation} 
where $W^{(0)} $, $\mathbf{b}^{(0)}$ are learnable parameters.
\paragraph{Decoder.} 
At the $t$-th decoding step, for the current partial solution $(x_1,\cdots, x_{t-1})^\intercal$, the inputs of the decoder $R^{(0)}$ and $\widetilde{H}^{(0)}$ can be expressed as
\begin{equation}
\label{Representative}
\begin{aligned}
R^{(0)}=\operatorname{Concat}(W_1\mathbf{h}_{x_1}, W_2\mathbf{h}_{x_{t-1}}), \ \ \ \ \ \ \ \\
\end{aligned}
\end{equation}
\begin{equation}
\begin{aligned}
\widetilde{H}^{(0)}=\operatorname{Concat}(W_1\mathbf{h}_{x_1}, W_2\mathbf{h}_{x_{t-1}},H_a),
\end{aligned}
\end{equation}
where $W_1$, $W_2 \in \mathbb{R}^{d \times d}$ are learnable matrices, $\mathbf{h}_{x_1}$, $\mathbf{h}_{x_{t-1}}$ refer to the destination node's embedding and the starting node's embedding, respectively. $H_a =\{ \mathbf{h}_i, i=1,\ldots,n, i \notin \{x_{1:t-1}\} \}$ denotes the set of all available node's embeddings.  

We select the starting and destination nodes as the \textbf{representative points} since they indicate the most crucial information: where the partial solution begins and ends. Accordingly, their embeddings are transformed into $R^{(0)} \in  \mathbb{R}^{2\times d}$, as illustrated in Equation (\ref{Representative}). 

Then the inputs $R^{(0)}$ and $\widetilde{H}^{(0)}$ are processed by the $L$ linear attention modules. For $l=1,\ldots,L$, the operations performed by the $l$-th linear attention module can be formalized as follows:  
\begin{equation}
\label{aggregate}
\begin{aligned}
    R^{(l)} & = \operatorname{AttentionLayer}(R^{(l-1)},\widetilde{H}^{(l-1)}),\\
\end{aligned}
\end{equation}
\begin{equation}
\label{broadcast}
\begin{aligned}
    \widetilde{H}^{(l-1)} & = \operatorname{AttentionLayer}(\widetilde{H}^{(l-1)}, R^{(l)} ), \quad \quad \ \\
\end{aligned}
\end{equation}
where $\widetilde{H}^{(l-1)},R^{(l-1)}$ represent the inputs, while $\widetilde{H}^{(l)},R^{(l)}$ denote the outputs. The term ``AttentionLayer'' is used to describe the classical attention layer as introduced by~\citep{vaswani2017attention}, more detailed information about it can be found in Appendix \ref{Attention Layer}.

The output of the final linear attention module is $\widetilde{H}^{(L)}$ and $ R^{(L)}$. Then, a linear projection and softmax function are applied to $ \widetilde{H}^{(L)}$, producing the selection probability of each available node. The starting and destination nodes are masked before the softmax calculation
\begin{equation}
\label{LEHD decoder equation}
\begin{aligned}
        u_i &= \begin{cases}
           W_O\widetilde{\mathbf{h}}_{i}^{(L)}, &\text{$i\neq1$ or $2$}\\
           -\infty, &\text{otherwise}
           \end{cases},\\
        \mathbf{p}^t &= \operatorname{softmax}(\mathbf{u}).
\end{aligned}
\end{equation}
where $W_O $ is a learnable matrix. The most suitable node, $x_t$, is selected based on $\mathbf{p}^t$. Finally, by calling the decoder $n$ times, a complete solution $\mathbf{x}=(x_{1},\ldots,x_{n})^\intercal$ is constructed.

\paragraph{Analysis.} Referring to Equation (\ref{aggregate}), the input dimensions are $2\times d$ for $R^{(l-1)}$ and $\widetilde{n}\times d$ for $\widetilde{H}^{(l-1)}$, where $\widetilde{n}$ represents the number of nodes fed into the decoder. Excluding constants such as $d$, the interaction in the attention layer between node embeddings in $R^{(l-1)}$ and $\widetilde{H}^{(l-1)}$ yields a computational complexity of $\mathcal{O}(\widetilde{n})$, linearly proportional to the number of input nodes. Similarly, Equation (\ref{broadcast}) exhibits the same linear computational complexity. The use of a specific number of representative points helps the model efficiently eliminate the need for explicit computations between all nodes during the attention process, thereby achieving linear computation and space complexity.

\section{Experiments}
We compare our proposed SIL method with other representative learning-based and classical solvers on TSP and CVRP instances with different scales and distributions.
\paragraph{Problem Setting.} 

We denote TSP and CVRP instances with 1,000 to 100,000 nodes as TSP/CVRP1K, 5K, 10K, 50K, and 100K, respectively. The TSP1K test set consists of 128 instances, and each of TSP5K, TSP10K, TSP50K, and TSP100K has 16 instances. Similarly, the CVRP test set includes the same number of instances, with capacities set to 250 for CVRP1K, 500 for CVRP5K, 1,000 for CVRP10K, and 2,000 for CVRP50K/100K. We generate the datasets following the standard generation procedure utilized in~\citep{kool2018attention}, except for TSP1K and TSP10K where we use the same test instances provided by~\citet{fu2021generalize} for fair comparison. The optimal solutions of TSP instances are computed using LKH3~\citep{LKH3}, while CVRP instances are solved via HGS~\citep{HGS}.

\paragraph{Model Setting.} 
For our proposed model, we set the embedding dimension ($d$) to 128, and the decoder employs $L=6$ linear attention modules, with each attention layer including 8 attention heads and a feed-forward layer with a hidden dimension of 512. 

\paragraph{Training.} 
As detailed in Appendix \ref{Warm Up Stage}, the model initially undergoes a warm-up training process on 20-node instances using the same RL-based method as in \citet{kwon2020pomo}, which takes about 15 hours. After that, we train the model on instances with sizes 1K and then leverage it as a base model to conduct separate training on larger scales, including 5K, 10K, 50K, and 100K. 

The training dataset sizes for scales 1K, 5K/10K and 50K/100K are 1,0000, 200, and 100, respectively. Each cycle of self-improved learning consists of 100 iterations of local construction and 20 epochs for model training. The training time varies depending on the scale and dataset size, ranging from one day to ten days. The Adam optimizer~\citep{Adam} is utilized for training the models, with an initial learning rate of 1e-4 and a decay rate of 0.97 per epoch. Throughout both the local reconstruction and training phases, the maximum length of the partial solution $l_{max}$ is maintained at 1000 to balance efficiency and effectiveness. The training procedure remains the same for TSP and CVRP. More detailed training settings, such as batch sizes, can be found in Appendix \ref{Hyperparameters of Training Procedure}. In all experiments, we use a single NVIDIA GeForce RTX 3090 GPU with 24GB of memory for both training and testing.

\paragraph{Baselines.}
We compare our method with \textbf{(1) Classical Solvers:} Concorde~\citep{applegate2006concorde}, LKH3~\citep{LKH3}; HGS~\citep{HGS}; \textbf{(2) Insertion Heuristic:} Random Insertion; \textbf{(3) Constructive Method:} POMO~\citep{kwon2020pomo}, and BQ~\citep{drakulic2023bq}, LEHD~\citep{luo2023neural} (4) \textbf{Two-stage Method:} Att-GCN+MCTS~\citep{fu2021generalize}, DIMES~\citep{qiu2022dimes}, DIFUSCO~\citep{sun2023difusco}, (5) \textbf{Decomposition-based Method:} L2D~\citep{li2021learning}, GLOP~\citep{ye2024glop}, TAM~\citep{hou2023generalize}, RBG~\citep{zong2022rbg}, NLNS~\citep{hottung2020neural}, L2I~\citep{lu2019learning}, and H-TSP~\citep{pan2023h-tsp}. 

\paragraph{Metrics and Inference.} 
For comparison, we provide the average objective value (Obj.), optimality gap (Gap), and inference time (Time) of each method. \textit{Obj.} indicates the solution length, with shorter being better. \textit{Gap} measures the solution difference from the ground truth (results produced by LKH for TSP and HGS for CVRP). \textit{Time}, recorded in seconds (s), minutes (m), or hours (h), reflects the efficiency in generating solutions for all test instances.

For most baseline methods, we execute their source code with default settings. The results of RBG, L2D, and NLNS are from~\citet{ye2024glop}, and the results of TAM, and L2I are from~\citet{hou2023generalize}. The results of Att-GCN+MCTS, DIMES, and DIFUSCO are from their original papers. For our method, a universal and fast random insertion method is employed for the initial solution generation, and then our model iteratively performs local reconstructions to improve the quality of the initial solution. We present the results of our Parallel local Re-Construction (PRC) method under various iterations.
\begin{table*}[t]
\centering
\caption{Comparative results on uniformly distributed TSP and CVRP instances. Asterisked (*) results are cited directly from original publications. ``N/A'' indicates the method exceeds the time limit (e.g., several days) or produces infeasible solutions. "OOM" indicates that the method exceeded memory limits.}
\resizebox{1\textwidth}{!}{
\begin{tabular}{ll | cc | cc | cc | cc | cc}
\toprule[0.5mm]
\multicolumn{2}{l|}{ }& \multicolumn{2}{c|}{TSP1K}   & \multicolumn{2}{c|}{TSP5K} & \multicolumn{2}{c|}{TSP10K}& \multicolumn{2}{c|}{TSP50K} & \multicolumn{2}{c}{TSP100K} \\ 
\multicolumn{2}{l|}{ Method} & Obj. (Gap) & Time  & Obj. (Gap) & Time & Obj. (Gap) & Time & Obj. (Gap) & Time & Obj. (Gap) & Time     \\ 
\midrule
\multicolumn{2}{l|}{LKH3}   & 23.12 (0.00\%) & 1.7m &  50.97 (0.00\%) & 12m & 71.78 (0.00\%) & 33m &  159.93 (0.00\%)  & 10h & 225.99 (0.00\%)  & 25h\\
\multicolumn{2}{l|}{Concorde} & 23.12 (0.00\%) & 1m &  50.95 (-0.05\%) & 31m & 72.00 (0.15\%) & 1.4h   & N/A  & N/A & N/A  & N/A \\
\multicolumn{2}{l|}{Random Insertion} & 26.11 (12.9\%) & \textless 1s & 58.06 (13.9\%) & \textless 1s & 81.82 (13.9\%) & \textless 1s  & 182.6 (14.2\%)  & 15.4s & 258.13 (14.2\%)  & 1.7m\\
\midrule
\multicolumn{2}{l|}{Att-GCN+MCTS*}  & 23.86 (3.20\%) & 6s &    $-$ & $-$ & 74.93 (4.39\%)  & 6.6m &  $-$ & $-$ & $-$  & $-$ \\

\multicolumn{2}{l|}{DIMES*}    & 23.69 (2.46\%) & 2.2m   & $-$ & $-$ & 74.06 (3.19\%) & 13m & $-$  & $-$ & $-$ & $-$ \\
\multicolumn{2}{l|}{DIFUSCO*}    & 23.39 (1.17\%) & 11.5s   & $-$ & $-$ & 73.62 (2.58\%) & 3.0m & $-$  & $-$ & $-$ & $-$ \\
\multicolumn{2}{l|}{H-TSP}  & 24.66 (6.66\%) & 48s   & 55.16 (8.21\%) & 1.2m & 77.75 (8.38\%) & 2.2m   &  \multicolumn{2}{c|}{OOM}  &   \multicolumn{2}{c}{OOM} \\
\multicolumn{2}{l|}{GLOP}    & 23.78 (2.85\%) & 10.2s  & 53.15 (4.26\%) & 1.0m & 75.04 (4.39\%) & 1.9m   &  168.09 (5.10\%) & 1.5m & 237.61 (5.14\%)  &  3.9m\\
\midrule
\multicolumn{2}{l|}{POMO aug$\times$8}    & 32.51 (40.6\%) & 4.1s & 87.72 (72.1\%)  & 8.6m  &  \multicolumn{2}{c|}{OOM} &  \multicolumn{2}{c|}{OOM}    &  \multicolumn{2}{c}{OOM} \\

\multicolumn{2}{l|}{BQ bs16}   & 23.43 (1.37\%) & 13s  & 58.27 (10.7\%) & 24s & \multicolumn{2}{c|}{OOM}  & \multicolumn{2}{c|}{OOM}   &   \multicolumn{2}{c}{OOM} \\

\multicolumn{2}{l|}{LEHD RRC1000}    &  \textbf{23.29 (0.72\%)} & 3.3m & 54.43 (6.79\%) & 8.6m & 80.90 (12.5\%) & 18.6m &  \multicolumn{2}{c|}{OOM}  &   \multicolumn{2}{c}{OOM} \\

\midrule
\multicolumn{2}{l|}{SIL PRC10} & 23.65 (2.31\%)  & 0.8s  & 52.68 (3.35\%) & 5.4s & 74.38 (3.46\%) & 5.7s & 167.64 (4.82\%) & 41s  & 237.68 (5.17\%) & 2.5m \\
\multicolumn{2}{l|}{SIL PRC50} & 23.46 (1.50\%)  & 4.2s  & 52.25 (2.51\%) & 22s & 73.77 (2.61\%) & 38s  &  165.85 (3.70\%) & 2.8m  & 235.06 (4.01\%) & 7.2m \\
\multicolumn{2}{l|}{SIL PRC100} & 23.39 (1.16\%)  & 8.1s & 52.12 (2.25\%) & 43s & 73.59 (2.36\%) & 1.4m  & 165.41 (3.43\%) & 6.9m  & 234.14 (3.61\%) & 13.1m \\
\multicolumn{2}{l|}{SIL PRC500} & 23.32 (0.87\%)  & 36s &  51.95 (1.92\%) & 3.9m & 73.37 (2.05\%) & 6.8m  & 164.72 (2.99\%) & 32m  & 233.00 (3.1\%) & 1.0h \\
\multicolumn{2}{l|}{SIL PRC1000} & 23.31 (0.81\%)  & 1.2m  & \textbf{51.92 (1.86\%)} & 7.6m & \textbf{73.32 (2.00\%)} & 13.7m  & \textbf{164.53 (2.87\%)} & 1.0h  & \textbf{232.66 (2.95\%)} & 2.0h \\
\bottomrule[0.5mm]
\toprule[0.5mm]
\multicolumn{2}{l|}{ }& \multicolumn{2}{c|}{CVRP1K} & \multicolumn{2}{c|}{CVRP5K} & \multicolumn{2}{c|}{CVRP10K}&  \multicolumn{2}{c|}{CVRP50K} & \multicolumn{2}{c}{CVRP100K}     \\ 
\multicolumn{2}{l|}{ Method} & Obj. (Gap) & Time  & Obj. (Gap) & Time & Obj. (Gap) & Time & Obj. (Gap) & Time & Obj. (Gap) & Time     \\ 
\midrule
\multicolumn{2}{l|}{HGS}      & 36.29 (0.00\%) & 2.5m  & 89.74 (0.00\%) & 2.0h & 107.40 (0.00\%) & 5.0h    & 267.73 (0.00\%)  & 8.1h & 476.11 (0.00\%)  & 24h\\
\multicolumn{2}{l|}{LKH3}    & 37.09 (2.21\%) & 3.3m   & 93.71 (5.19\%)  & 1.33h  & 118.76 (10.6\%) &  1.74h   & 399.12 (49.1\%)  & 15.8h & N/A  & N/A \\
\multicolumn{2}{l|}{Random Insertion}    & 57.42 (58.2\%) & \textless 1s  
 & 154.4 (72.0\%) & \textless 1s & 191.80 (78.6\%) & \textless 1s  & 490.56 (83.2\%)  & \textless 1s & 943.87 (98.3\%)  & 2s\\
\midrule
\multicolumn{2}{l|}{GLOP-G (LKH3)}    & 39.50 (8.83\%) & 1.3s & 98.90 (10.2\%)  & 6.8s & 116.28 (8.27\%)  & 11.2s&  \multicolumn{2}{c|}{OOM}  &    \multicolumn{2}{c}{OOM} \\
\midrule
\multicolumn{2}{l|}{POMO aug$\times$8}   & 84.89 (134\%) & 4.8s  & 393.3 (338\%) & 11m & \multicolumn{2}{c|}{OOM} &  \multicolumn{2}{c|}{OOM}  &  \multicolumn{2}{c}{OOM} \\
\multicolumn{2}{l|}{BQ bs16}   & 38.17 (5.17\%) & 14s  & 104.4 (16.3\%) & 2.6m & \multicolumn{2}{c|}{OOM}  & \multicolumn{2}{c|}{OOM}  &  \multicolumn{2}{c}{OOM}  \\
\multicolumn{2}{l|}{LEHD RRC1000}    & 37.43 (3.15\%) & 3.4m  & 101.1 (12.6\%) & 31m & 138.73 (29.2\%) & 41m  & \multicolumn{2}{c|}{OOM}   &  \multicolumn{2}{c}{OOM} \\

\midrule
\multicolumn{2}{l|}{SIL PRC10  } & 37.88 (4.37\%) & 0.6s & 93.87 (4.60\%)  & 3.5s & 111.73 (4.03\%)  & 5.5s & 278.92 (4.18\%)  & 21.5s & 495.96 (4.17\%) & 40s  \\
\multicolumn{2}{l|}{SIL PRC50  } & 37.33 (2.88\%) & 2.9s  & 91.58 (2.05\%)  & 17s & 108.68 (1.19\%)  & 28s  & 267.88 (0.06\%)  & 1.7m & 473.65 (-0.52\%) & 3.0m \\
\multicolumn{2}{l|}{SIL PRC100  } & 37.16 (2.41\%) & 6.6s  & 91.05 (1.46\%)  & 40s & 107.53 (0.12\%)  & 1.1m   & 264.67 (-1.14\%)  & 4.4m & 467.60 (-1.79\%) & 7.6m \\
\multicolumn{2}{l|}{SIL PRC500  } & 36.90 (1.68\%) & 38s  & 90.24 (0.56\%)  & 3.8m & 106.34 (-0.99\%)  & 6.3m  & 261.16 (-2.45\%)  & 24m & 460.66 (-3.25\%) & 44m \\
\multicolumn{2}{l|}{SIL PRC1000  } & \textbf{36.83 (1.49\%)} & 1.3m  &   \textbf{90.04 (0.34\%)}& 7.6m & \textbf{106.07 (-1.24\%)}  & 13m  & \textbf{260.43 (-2.72\%)}  & 49m& \textbf{459.38 (-3.51\%}) & 1.5h \\
\bottomrule[0.5mm]
\end{tabular}
}
\label{table1-uniform}
\end{table*}

\begin{table}[htbp]
\centering
\caption{Comparative results on CVRP instances with different capacity settings.}
\label{CVRPdata_2}
\resizebox{1\columnwidth}{!}{
\begin{tabular}{ll | c | c | c | c }
\toprule[0.5mm]
\multicolumn{2}{l|}{ }& CVRP1K  & CVRP2K &  CVRP5K& CVRP7K \\ 
\multicolumn{2}{l|}{ Method}
& Obj. (Time)  & Obj. (Time) & Obj. (Time) & Obj. (Time)     \\ 
\midrule
\multicolumn{2}{l|}{HGS}  & 41.2 (5m) & 57.2 (5m) & 126.2  (5m) & 172.1 (5m) \\
\multicolumn{2}{l|}{LKH3}   & 46.4 (6.2s) & 64.9 (20s) & 175.7 (2.5m) & 245.0 (8.4m) \\
\multicolumn{2}{l|}{Random Insertion}   & 66.3 (\textless 1s) & 95.3 (\textless 1s) & 225.4 (\textless 1s) & 309.2 (\textless 1s) \\
\midrule
\multicolumn{2}{l|}{L2I}  & 93.2 (6.3s) & 138.8 (25s) & $-$ & $-$  \\
\multicolumn{2}{l|}{NLNS}  &53.5 (3.3m) & $-$  &  $-$ & $-$\\
\multicolumn{2}{l|}{L2D}  & 46.3 (1.5s) & 65.2 (38s) & $-$  & $-$ \\
\multicolumn{2}{l|}{RBG}  & 74.0 (13s) & 138  (42s) & $-$ & $-$ \\
\multicolumn{2}{l|}{TAM-LKH3} & 46.3 (1.8s) & 64.8  (5.6s) & 144.6 (17s) & 196.9 (33s) \\
\multicolumn{2}{l|}{TAM-HGS}  & $-$  & $-$  & 142.8 (30s) & 193.6 (52s) \\
\multicolumn{2}{l|}{GLOP-G}  & 47.1 (0.4s) & 63.5 (1.2s) & 141.9 (1.7s) & 191.7 (2.4s) \\
\multicolumn{2}{l|}{GLOP-G (LKH-3)}  & 45.9 (1.1s) & 63.0 (1.5s) & 140.6 (4.0s)& 191.2 (5.8s)\\
\midrule
\multicolumn{2}{l|}{POMO aug$\times$8}  & 101 (4.6s) & 255 (51s) & 632.9 (11m)& OOM \\
\multicolumn{2}{l|}{BQ bs16}  & 43.1 (14s) & 60.9 (33s) & 136.4 (2.4m)& 186.8 (5.7m) \\
\multicolumn{2}{l|}{LEHD RRC1000}  & 42.4 (3.4m) & 59.4 (4.6m) & 132.7 (10m)& 180.6 (19m) \\
\midrule
\multicolumn{2}{l|}{SIL PRC10}  & 43.4 (0.6s) & 59.0 (1.2s) & 128.0 (2.7s)  & 171.7 (4.0s) \\
\multicolumn{2}{l|}{SIL PRC50}  & 42.7 (3.0s) & 58.0 (5.4s) & 125.3 (13s) & 168.4 (19s) \\
\multicolumn{2}{l|}{SIL PRC100}  & 42.5 (7.0s)  & 57.7 (13s) & 124.5 (31s) & 167.3 (44s) \\
\multicolumn{2}{l|}{SIL PRC500}  & 42.1 (40s) & 57.2 (1.2m) & 123.4 (3.0m) & 165.8 (4.2m) \\
\multicolumn{2}{l|}{SIL PRC1000}  & \textbf{42.0} (1.3m) & \textbf{57.1} (2.4m) &  \textbf{123.1} (5.9m) & \textbf{165.3} (8.3m) \\
\bottomrule[0.5mm]
\end{tabular}
}
\end{table}

\begin{table}[htbp]
\centering
\caption{Empirical results on TSPLib and  CVRPLib. BKS refers to the ``Best Known Solution".}
\resizebox{0.99\columnwidth}{!}{
\begin{tabular}{ll | c| c  | c }
\toprule[0.5mm]
   \multicolumn{2}{l|}{  } & \multicolumn{3}{c}{TSPLib }\\
 \midrule
 \multicolumn{2}{l|}{Scale (\#)} & 3K\textless $N\leq$ 10K (4)& 10K\textless $N\leq$ 20K (4) & Total (8)\\
  \multicolumn{2}{l|}{  }& Gap (Time)   &  Gap (Time)   & Gap (Time)  \\
\midrule
\multicolumn{2}{l|}{BKS}   & 0.00\% ($-$) &  0.00\% ($-$) &  0.00\% ($-$) \\
\multicolumn{2}{l|}{LKH3}   & 0.08\% (6.6m) &  0.06\% (1.0h) &  0.07\% (35m) \\
\midrule
\multicolumn{2}{l|}{GLOP}   & 8.10\% (15s) &  5.87\% (52s) &  6.99\% (34s) \\
\multicolumn{2}{l|}{BQ bs16}   & 17.6\% (2.2m) &  OOM &  17.6\% (2.2m) \\
\multicolumn{2}{l|}{LEHD RRC1000}   & 10.4\% (25m)& 19.3\% (55m) &  14.2\% (38m) \\
\midrule
\multicolumn{2}{l|}{SIL PRC10}   & 5.68\% (27s) &  5.49\% (28s) &  5.59\% (27s) \\
\multicolumn{2}{l|}{SIL PRC100}   & 3.50\% (4.1m) &  3.78\% (4.8m) &  3.64\% (4.5m) \\
\multicolumn{2}{l|}{SIL PRC1000}   & \textbf{2.92\%} (41m) &  \textbf{3.14\%} (49m) &  \textbf{3.03\%} (45m) \\

\midrule

  \multicolumn{2}{l|}{  } & \multicolumn{3}{c}{CVRPLib }\\
 \midrule
 \multicolumn{2}{l|}{Scale (\#) } & 6K\textless $N\leq$ 11K (4)& 15K\textless $N\leq$ 30K (4)& Total (8)\\
 \multicolumn{2}{l|}{  }& Gap (Time)   &  Gap (Time)   & Gap (Time)  \\
\midrule
\multicolumn{2}{l|}{BKS}   & 0.00\% ($-$) &  0.00\% ($-$) &  0.00\% ($-$) \\
\multicolumn{2}{l|}{HGS}   & 3.31\% (5h)&  6.99\% (5h)& 5.15\% (5h)\\
\multicolumn{2}{l|}{LKH3}   & 13.6\% (2.1h)&  $-$ & 13.6\% (2.1h)\\
\midrule
\multicolumn{2}{l|}{TAM-LKH3}   & 24.9\% (37s) & 32.1\% (3.0m) &  27.3\% (1.4m)\\
\multicolumn{2}{l|}{GLOP-G (LKH3)}   & 17.9\% (8.5s) &  23.3\% (1.2m) &  20.6\% (39s) \\
\multicolumn{2}{l|}{LEHD RRC1000}   & 16.9\% (40m) &  OOM &  16.9\% (40m) \\
\midrule
\multicolumn{2}{l|}{SIL PRC10}   & 9.78\% (23s) & 14.1\% (27m) &  11.9\% (25s)\\
\multicolumn{2}{l|}{SIL PRC100}   & 7.31\% (4.3m) & 11.0\% (5.3m) &  9.14\% (4.8m)\\
\multicolumn{2}{l|}{SIL PRC1000}   & \textbf{6.02\%} (48m) &  \textbf{9.36\%} (59m) &  \textbf{7.69\%} (54m)\\
\bottomrule[0.5mm]
\end{tabular}
}
\label{Performance on TSPLib and CVRPLib}
\vskip -0.1in
\end{table}

\subsection{Experimental Results}
Table \ref{table1-uniform} presents the results on uniformly distributed TSP and CVRP instances. From the results, we can observe that our method is consistently demonstrating superior performance. For TSP, our method with only 100 PRC iterations outperforms the other learning-based methods in terms of both solution quality and solving efficiency on all problems except for TSP1K. On TSP1K, our method is slightly worse than LEHD RRC1000 but consumes less solving time. By increasing the PRC iteration to 500 and 1000, the performance of our method can be further improved. Overall, our method shows good scalability and can achieve outstanding performance even for very large-scale problem instances. For the CVRP instances, as shown in Table \ref{table1-uniform}, our method with 50 PRC iterations can surpass all the other learning-based methods on all scales. Remarkably, our method outperforms the classical solver HGS at CVRP10K to CVRP100K, which can be regarded as a notable achievement for constructive NCO. 

Table \ref{CVRPdata_2} shows the results obtained by each method on CVRP instances with different capacity settings. For our method, the model is trained on CVRP1K instances and subsequently tested across larger instances, including CVRP2K, CVRP5K, and CVRP7K. Even with a small number of PRC iterations, our method can achieve better objective values than the other learning-based methods, indicating good scalability and efficiency of SIL in solving larger and more complex CVRP instances. The experimental results obtained by each method on large-scale real-world TSPLib~\cite{Rein91} and CVRPLib~\cite{uchoa2017new} instances are provided in Table \ref{Performance on TSPLib and CVRPLib}. It is obvious that our method significantly outperforms TAM-LKH3, GLOP, BQ, and LEHD, even with only 10 PRC iterations.

\paragraph{Training Progress for CVRP with 100K Nodes.}

\begin{figure}[t]
\centering
\includegraphics[width=1\columnwidth,height=5.0cm]{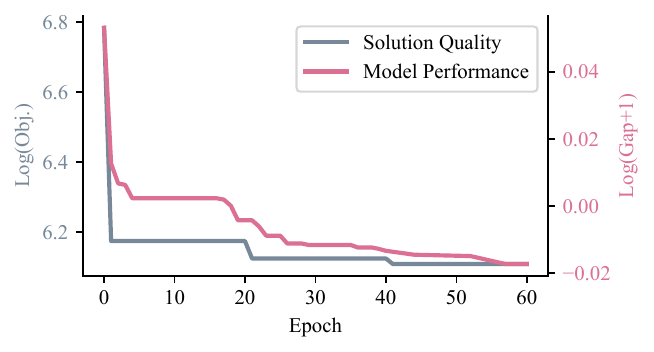}
\caption{SIL training progress on CVRP with $100,000$ nodes.}
\label{Improving Process on CVRP10W}
\end{figure}

In Figure~\ref{Improving Process on CVRP10W}, we show the changing curves with respect to the solution quality and the model performance. During the training, there is a consistent enhancement in the solution quality, leading to improved model performance. The training process terminates when the solution quality and model performance converge. Note that SIL can outperform HGS on CVRP100K after roughly 20 epochs.

\begin{table}[t]
\centering
\caption{Comparisons of peak memory usage over LEHD and our proposed linear attention model on TSP instances with different scales.}
\resizebox{1\columnwidth}{!}{
\begin{tabular}{l|ccccc}
\toprule[0.5mm]
\multicolumn{1}{c|}{} & \multicolumn{5}{c}{Peak Memory Usage (MB) /Instance }    \\
\multicolumn{1}{c|}{Method} & TSP1K  & TSP5K   & TSP10K    & TSP50K  & TSP100K   \\ \hline
LEHD              & 101.1   & 2317.6  & 9202.8   & OOM & OOM  \\
Ours              & 36.7   & 51.7  & 94.3   & 426.4 & 844.0  \\
\bottomrule[0.5mm]     
\end{tabular}
}
\label{Memory Usage Comparison}
\end{table}

\begin{figure}[t]
\centering
\includegraphics[width=0.99\columnwidth,height=5.1cm]{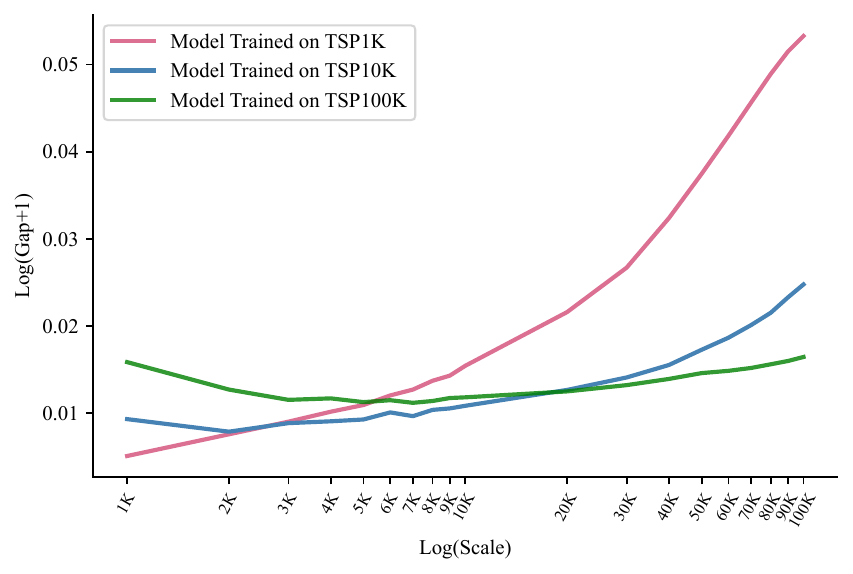}
\caption{Comparision of models trained on different scales.}
\label{range_test_TSP}
\end{figure}

\paragraph{Memory Usage: Linear Attention Model V.S. Quadratic Attention Model.}
In Table \ref{Memory Usage Comparison}, we present a comparison of peak memory usage between our model and LEHD. It can be found that our linear attention model has a significantly lower memory usage in contrast to LEHD. A more comprehensive discussion on memory usage can be found in Appendix \ref{Memory Consumption of the Proposed Linear Attention}.

\paragraph{Comparision of Models Trained on Different Scales.}
To illustrate the generalization ability of each trained model over different scales, we use 20 TSP datasets whose scales are from 1K to 100K to assess each model's performance. Each dataset contains 16 instances, and we conduct each model with 100 PRC iterations on each instance and plot the results in Figure \ref{range_test_TSP}. The result reveals that each model can maintain the best performance on the datasets that have similar scales to its own training dataset. Their performance degrades on datasets that are very different from the training dataset in scales. Overall, the model trained on large-scale instances is more robust than those trained on small-scale ones.

\section{Conclusion and Future Work}
In this paper, we have proposed a scalable NCO method called SIL, which enables the model to be directly trained on large-scale instances without any labeled data. SIL adopts a local reconstruction strategy to produce enhanced solutions as pseudo-labels to guide model training; while the improved model in turn can boost the local reconstruction to generate better solutions. Through iterative loops, SIL can continuously improve the performance of the model. Moreover, we have developed a linear complexity attention mechanism, which significantly reduces the computational cost in both training and inference, making it more suitable for solving large-scale CO problems. Extensive experimental results on TSP and CVRP with up to 100K nodes in both uniform and real-world distributions demonstrate the superior performance of SIL.

Although our method achieves good overall performance, it still cannot outperform classical solvers on certain problems, such as LKH3 for TSPs. To narrow this gap further, we plan to design a more powerful local reconstruction method to enable more efficient self-improved learning in the future. In addition, we will investigate more advanced strategies to accelerate the training progress in the future.

\clearpage

\section*{Impact Statement}

This paper presents work whose goal is to advance the field of Machine Learning. There are many potential societal consequences of our work, none which we feel must be specifically highlighted here.

\bibliography{references}
\bibliographystyle{icml2024}

\newpage
\appendix

\section{Warm Up Stage}
\label{Warm Up Stage}

The model first employs the RL-based training method~\citep{kwon2020pomo} to train on small-scale instances. 

Given an instance $\mathbf{s}$ with $n$ nodes, the model parameterized by $\boldsymbol{\theta}$ begins with designating $n$ different nodes $\left\{a^1, a^2, \ldots, a^n\right\}$ as starting nodes and simultaneously generates a set of solutions $\left\{\boldsymbol{\tau}^1, \boldsymbol{\tau}^2, \ldots, \boldsymbol{\tau}^n\right\}$. The return $R(\boldsymbol{\tau}^i)$ of each solution $\boldsymbol{\tau}^i$ is defined as its length. To maximize the expected return $J$, we use gradient ascent with an approximation
\begin{equation}
    \nabla_{\boldsymbol{\theta}} J(\boldsymbol{\theta}) \approx \frac{1}{n} \sum_{i=1}^n\left(R\left(\boldsymbol{\tau}^i\right)-b(s)\right) \nabla_{\boldsymbol{\theta}} \log p_{\boldsymbol{\theta}}\left(\boldsymbol{\tau}^i \mid s\right).
\end{equation}

where $p_{\boldsymbol{\theta}}\left(\boldsymbol{\tau}^i \mid s\right) = \prod_{t=2}^n p_{\boldsymbol{\theta}} \left(a_t^i \mid s, a_{1: t-1}^i\right)$, and $a_{1: t-1}^i$ denotes the partial solution at step $t$,  $p_{\boldsymbol{\theta}} \left(a_t^i \mid s, a_{1: t-1}^i\right)$ denotes the probability of selecting node $a_t^i$ at setp $t$ conditions on the instance and the partial solution. $b_i(s)$ is the baseline defined as $b(s)=\frac{1}{n} \sum_{i=1}^n R\left(\tau^i\right)$. 

In each training iteration, a batch of instances is randomly generated, and the corresponding gradients are calculated and utilized to optimize the model. This warm-up stage enables the model to perform the local reconstruction on large-scale problem solutions and produce enhanced solutions for further training.

In the warm-up stage, for both training on the TSP and CVRP set, policy gradients are averaged from a batch of 64 instances. Adam optimizer~\citep{Adam} is used with a learning rate of 10-4. We define one epoch of 50,000 training instances generated randomly on the fly. The training settings can be found in Table \ref{tab: warm-up hyperparameters}.

\begin{table}[htbp]
\centering
\caption{Hyperparameters and the training time in the warm-up phase.}
\resizebox{0.25\textwidth}{!}{
\begin{tabular}{l|cc}
\toprule[0.5mm]
       & TSP20  & CVRP20  \\ 
\hline
epoch                  & 100      & 100    \\
batch size             & 64      & 64    \\
learning rate          & 1e-4   & 1e-4  \\
Training time          & 14.62h   & 15.42h  \\
\bottomrule[0.5mm]     
\end{tabular}
}
\label{tab: warm-up hyperparameters}
\end{table}

\section{Generate the Initial Solution}
\label{Generate the Initial Solution}
The initial solution of the given instance can be generated by the simple heuristics method like random insertion or the model's greedy search. 

\paragraph{Random insertion.} By this method, an incomplete solution is extended to a complete one by inserting random nodes one node at a time. When node $i$ is inserted, the place of insertion (between adjacent nodes $j$ and $k$ in the tour) is selected such that it minimizes the insertion costs $d_{ji}+d_{ik}-d_{jk}$, where $d_{ji}$, $d_{ik}$ and $d_{jk}$ represent the distances from node $j$ to $i$, $i$ to $k$ and $j$ to $k$, respectively.

\begin{figure*}[t]
\centering
\includegraphics[width=1.99\columnwidth]{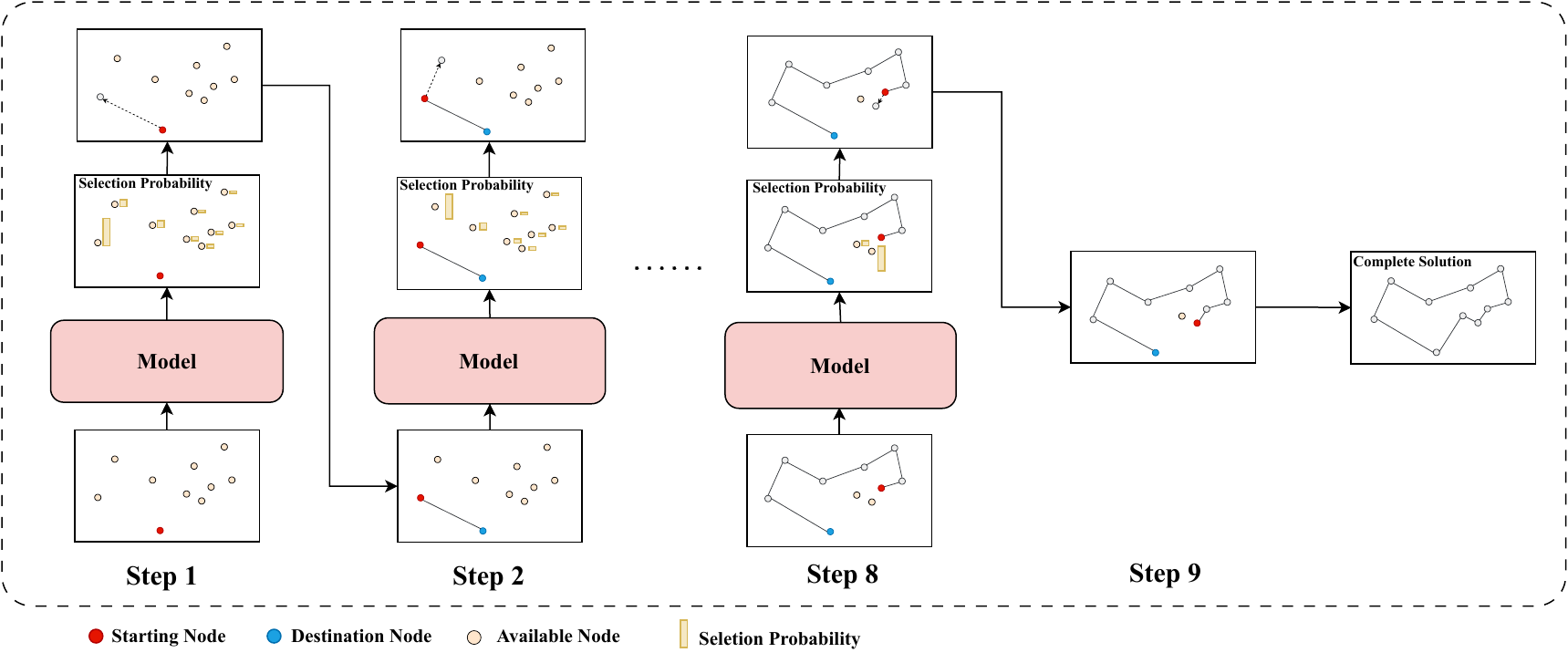}
\caption{The solution construction process.}
\label{Solution construction process}
\end{figure*}

\paragraph{Greedy search performed by the model.} As shown in Figure \ref{Solution construction process}, in the initial step, the model randomly chooses a node, leaving the others as available nodes. In each subsequent step, the model assigns a selection probability to each available node and chooses the one with the highest probability to extend the incomplete solution. The selected node is then marked as unavailable. This process continues until no available nodes remain, resulting in the generation of a complete solution.

\section{Memory Consumption of the Proposed Linear Attention}
\label{Memory Consumption of the Proposed Linear Attention}
In Table \ref{Memory Usage Comparison} within the main body of the paper, we conduct a comparative analysis of memory usage between the overall LEHD model and our newly proposed SILA model across various instance sizes. The LEHD model utilizes a self-attention mechanism in its architecture, while our model employs our innovative linear attention module. It's crucial to recognize that the overall memory usage of a model is not solely dependent on its attention mechanism; other components, such as linear projection, also contribute. To accurately assess the impact of the attention mechanisms on memory consumption, we performed dedicated tests. The results, detailed in Table \ref{Memory Usage Comparison of attention mechanism}, consistently show minimal memory usage for our linear attention module across all sizes of instances. This indicates that in our model, the linear projection's memory usage is predominant. In contrast, the self-attention mechanism is the principal consumer of memory in the LEHD model at every scale. This contrast highlights the efficiency of our linear attention approach in lowering computational complexity.
\begin{table}[htbp]
\centering
\caption{Comparisons of peak memory usage over self-attention and our proposed linear attention on TSP instances with different scales. We use the command $\textit{torch.cuda.max\_memory\_allocated()}$ to check the peak memory usage.}
\resizebox{1.0\columnwidth}{!}{
\begin{tabular}{l|ccccc}
\toprule[0.5mm]
\multicolumn{1}{c|}{} & \multicolumn{5}{c}{Peak Memory Usage of Attention Mechanism (MB)/Instance}    \\
\multicolumn{1}{c|}{Method} & TSP1K  & TSP5K   & TSP10K    & TSP50K  & TSP100K   \\ \hline
Self-Attention              & 78.2   & 2287.3  & 9165.8   & OOM & OOM  \\
Linear Attention              & \textless1   & \textless1  & 1.7   & 8.4 & 16.8  \\
\bottomrule[0.5mm]     
\end{tabular}
}
\label{Memory Usage Comparison of attention mechanism}
\end{table}

\begin{table*}[htbp]
\centering
\caption{Comparative results on real-world large-scale TSPLib and CVRPLib instances. \textit{BKS} refers to the ``Best Known Solution".}
\label{Performance on TSPLib and CVRPLib full}
\resizebox{0.99\textwidth}{!}{
\begin{tabular}{l | cc | cc | cc | cc |  cc |  cc |  cc |  cc }
\toprule[0.5mm]
  TSPLib Instance & \multicolumn{2}{c|}{pcb3038}  & \multicolumn{2}{c|}{fnl4461} & \multicolumn{2}{c|}{rl5915} & \multicolumn{2}{c|}{rl5934} & \multicolumn{2}{c|}{rl11849}    & \multicolumn{2}{c|}{brd14051}& \multicolumn{2}{c|}{d15112} & \multicolumn{2}{c}{d18512}  \\
Scale  &  \multicolumn{2}{c|}{3038} & \multicolumn{2}{c|}{4461} & \multicolumn{2}{c|}{5915} & \multicolumn{2}{c|}{5934} & \multicolumn{2}{c|}{11849}   & \multicolumn{2}{c|}{14051}  & \multicolumn{2}{c|}{15112}  & \multicolumn{2}{c}{18512}    \\
  & Obj. & Time &  Obj. & Time    &   Obj. & Time   & Obj. & Time  &  Obj. & Time   &  Obj. & Time   &   Obj. & Time   &   Obj. & Time   \\
\midrule

BKS & 137694 & $-$ & 182566  & $-$ & 565530 & $-$ & 556045 &$-$ & 923288 & $-$   & 469385 & $-$ & 1573084 & $-$ & 645238 & $-$ \\
\midrule
 LKH3   & 137779 & 3.5m  & 182798 & 4m & 565590 & 9m  &  556589 & 10m  & 923420 & 29m &   469792 & 56m &  1573426  & 1.3h & 645914  & 1.4h \\
 \midrule
 GLOP    &  144863 & 9s  & 190711 & 14s & 625171 & 19s  & 623892  & 19s & 1006256  & 39s  & 491759 & 47s & 1648338  & 53s  & 677051 & 1.1m \\
 BQ bs16    &   153546 & 1m  & 209363 & 1.8m & 676264 & 3m   & 692434  & 3m & \multicolumn{2}{c|}{OOM}     &    \multicolumn{2}{c|}{OOM}    &  \multicolumn{2}{c|}{OOM} &  \multicolumn{2}{c}{OOM}    \\
 LEHD RRC1000  &  147436 & 24m   & 202439 & 25m & 630378 & 26m  &  623893 & 26m & 1088125 & 44m  & 568829 & 57m & 1869288  & 1.1h &    \multicolumn{2}{c}{OOM}  \\
 \midrule
 SIL PRC10 & 142355 & 29s & 189552 & 32s  & 611945  & 17s & 596565 & 29s & 999008  & 22s & 492147 & 27s &  1640123 & 27s & 675330   & 37s \\
  SIL PRC50 & 141603 & 2.1m & 187295 & 2.0m  & 592000  & 1.8m & 586138 & 2.3m & 977325  & 2.2m & 487180 & 2.2m &  1623307 & 2.5m & 668100   & 2.9m \\
SIL PRC100 & 141080 & 4.1m & 186970 & 4.2m  & 588711  & 3.6m & 584051 & 4.5m & 972686  & 4.5m & 486452 & 4.5m &  1619784 & 4.9m & 665683   & 5.5m \\
SIL PRC500 & 140607 & 20.9m & 186008 & 20.0m  & 586990  & 20.1m & 581172 & 21.0m & 966077  & 22.7m & 483831 & 23.0m &  1613974 & 25.1m & 662878   & 25.8m \\
SIL PRC1000 & 140408 & 41.8m & 185934 & 40.7m  & 586602  & 41.2m & 579109 & 40.8m & 964661  & 45.6m & 483169 & 47.3m &  1612522 & 49.6m & 662178   & 51.5m \\
\bottomrule[0.5mm]
\toprule[0.5mm]

  CVRPLib Instance & \multicolumn{2}{c|}{Antwerp1}  & \multicolumn{2}{c|}{Antwerp2} & \multicolumn{2}{c|}{Ghent1} & \multicolumn{2}{c|}{Ghent2} & \multicolumn{2}{c|}{Brussels1}& \multicolumn{2}{c|}{Brussels2} & \multicolumn{2}{c|}{Flanders1} & \multicolumn{2}{c}{Flanders2}   \\
Scale  &  \multicolumn{2}{c|}{6000}  & \multicolumn{2}{c|}{7000} & 
  \multicolumn{2}{c|}{10000} & \multicolumn{2}{c|}{11000} & \multicolumn{2}{c|}{15000}  & \multicolumn{2}{c|}{16000}  & \multicolumn{2}{c|}{20000} & \multicolumn{2}{c}{30000} \\
  & Obj. & Time &  Obj. & Time    &  Obj. & Time  & Obj. & Time  &  Obj. & Time   &  Obj. & Time   &   Obj. & Time   &   Obj. & Time  \\
\midrule
BKS & 477277 & $-$ & 291350 & $-$  & 469531  & $-$ & 257748 & $-$ &  501719 & $-$ & 345468  & $-$ & 7240118  & $-$ & 4373244 &  $-$  \\
\midrule
 HGS  & 484516 & 5h & 302035 & 5h  & 483599  & 5h & 270787 & 5h & 524082  & 5h & 367129 & 5h & 7493715  & 5h & 4973217  &  5h  \\
 LKH3    & 493904 & 1.2h & 312353 &  4.1h & 558776  & 2.1h & 320891 & 1h & $-$  & $-$   & $-$  & $-$  & $-$   & $-$  & $-$   & $-$    \\
 \midrule
TAM-LKH3     & 591871 & 25s & 357049 & 32s &  608184 & 37s &  318705 & 56s &  637936 & 2.8m & 473533 & 3.1m & $-$  &  $-$ & $-$ & $-$ \\
 GLOP-LKH3     & 568344 & 9s & 342963 & 5s &  555379 & 13s &  300145 & 7s &  633039 & 17s & 406132 & 12s & 8978964  &  24s & 5484725 & 3.8m \\
 LEHD RRC1000     & 519779 & 33m & 336290 & 36m  &  550660 & 43m & 324172 & 49m &   \multicolumn{2}{c|}{OOM}  &  \multicolumn{2}{c|}{OOM}  &   \multicolumn{2}{c|}{OOM}  &   \multicolumn{2}{c}{OOM}    \\
 \midrule
 SIL PRC10  & 521285 & 23s & 318087 & 22s  & 509167  & 23s & 289403 & 23s & 552879  & 25s & 392716 & 25s & 8050546  & 28s &  5304990  & 32s \\
 SIL PRC50  & 513269 & 1.8m & 311621 & 1.8m  & 505264  & 1.9m & 282908 & 1.9m & 545862  & 2.1m & 383406 & 2.1m & 7945793  & 2.3m &  5113314  & 2.6m \\
SIL PRC100  & 511670 & 4.3m & 309448 & 4.2m  & 502978  & 4.4m & 280187 & 4.4m & 543058  & 4.8m & 380948 & 4.8m & 7907869  & 5.3m &  5078369  & 6.1m \\
SIL PRC500  & 507815 & 24m & 306076 & 23m  & 499989  & 25m & 276199 & 25m & 539105  & 27m & 375853 & 27m & 7841531  & 30m &  5010814  & 34m \\
SIL PRC1000  & 506853 & 47m & 305227 & 47m  & 498757  & 49m & 275540 & 49m & 537831  & 54m & 373845 & 54m & 7818513  & 1.9h &  4986724  & 1.1h \\
\bottomrule[0.5mm]
\end{tabular}
}
\end{table*}

\section{Detailed Results on TSPLib and CVRPLib.}
We select the large-scale instance in TSPLib and CVRPLib for testing and algorithm comparison. We present more detailed results in Table \ref{Performance on TSPLib and CVRPLib full}.

\section{Available Nodes' Selection Probabilities V.S. Distance to Starting Node}
\label{Selection Probabilities V.S. Distance}
\begin{figure}[H]
\centering
\includegraphics[width=0.8\columnwidth]{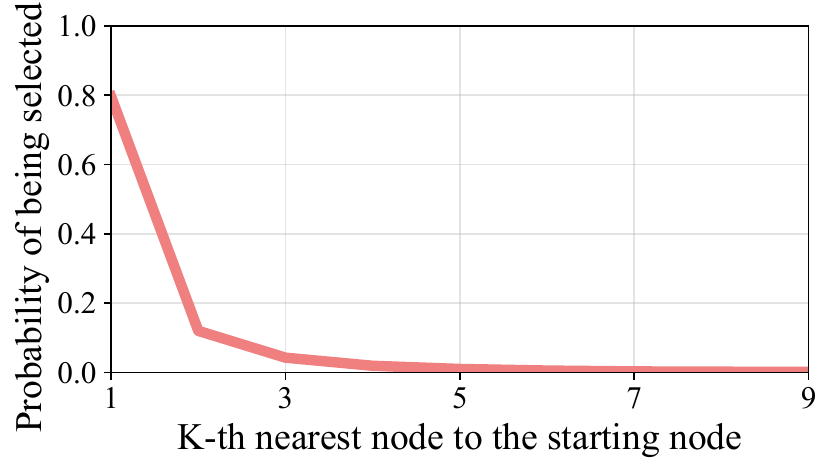}
\caption{The available nodes' probabilities of being selected relative to their distance to the starting nodes for TSP1000.}
\label{prob_distance}
\end{figure}
The decoder sequentially constructs the solution by adding one node at a time over $n$ steps. 
At each step, the decoder employs self-attention to capture the dynamic relations of the starting, destination, and available nodes, assigning a selection probability to each available node. Then it selects and adds one of them to the partial solution. The execution of self-attention with $\mathcal{O}(N^2)$ time and space complexity per step renders the decoding process computationally intensive, particularly for large-scale instances.

However, the probabilities assigned to nodes are not uniformly distributed and indeed extremely biased toward the starting node. To analyze this pattern, we employ the LEHD model~\citep{luo2023neural} to construct solutions for $128$ TSP1K instances and then record all available nodes' selection probability concerning its proximity to the starting node at each step, as shown in Figure \ref{prob_distance}. The findings reveal that nodes in closer vicinity to the starting node exhibit significantly higher selection probabilities, which drop sharply for those further away.

The distant nodes' explicit involvement in the self-attention mechanism may be unnecessary since they have almost no chance of being selected at each step. However, they may offer useful information about the distribution and structure of the instance graph, so they also should not be directly removed from the self-attention. Considering this, one possible strategy is to implicitly engage distant nodes in self-attention. To this end, we propose a novel linear attention module.

\section{Model Details}
\label{Model Details}
We employ the LEHD model~\citep{luo2023neural} as the backbone model and integrate the proposed linear attention module into it. We remove the attention layer in the encoder.

\subsection{Attention Layer}
\label{Attention Layer}
The attention layer comprises two sub-layers: the multi-head attention (MHA) sub-layer and the feed-forward (FF) sub-layer \citep{vaswani2017attention}. The normalization is removed from our model to enhance scalability performance following~\citet{luo2023neural}. 

With $X^{(l-1)} \in R^{N\times d}$ and $C^{(l-1)} \in R^{M\times d}$ as input, the operations performed by the $l$-th ``$\operatorname{Attention layer}$" in Equation (\ref{aggregate}) and (\ref{broadcast}) can be formalized as follows respectively:

\begin{equation}
\label{aggregate1}
\begin{aligned}
    X^{(l-1)\prime} & = \operatorname{MHA}(X^{(l-1)},C^{(l-1)})+X^{(l-1)},\\
    X^{(l)} & = \operatorname{FF}( X^{(l-1)\prime})+ X^{(l-1)\prime},\\
\end{aligned}
\end{equation}

\paragraph{multi-head attention.} The term``$\operatorname{MHA}$'' is the multi-head attention mechanism, which can be formalized as:
\begin{equation}
\begin{aligned}
    &\operatorname{MHA}(X^{(l-1)},C^{(l-1)})=\operatorname{Concat}(O_1,\ldots,O_h)W_O,\\
    &\quad \text{where}\quad O_i=\operatorname{Attention}_i(X^{(l-1)},C^{(l-1)}), \\ 
    & \quad \quad \quad \quad \quad \quad \quad \quad \quad \quad \quad   \forall i\in\{1,\ldots,h\},
\end{aligned}
\label{mha}
\end{equation}
$W_O \in \mathbb{R}^{d\times d}$ is a learnable matrix. $h$ is the head number, typically set to 8. 

\paragraph{Feed-forward layer.} The term ``$\operatorname{FF}$'' refers to the feed-forward layer, which consists of two linear transformations with a ReLU activation in between, and can be formalized as:
\begin{equation}
    \quad \text{FF}(X^{(l-1)})=\left(\text{ReLU}(X^{(l-1)}W_{1}+\mathbf{b}_{1})\right)W_{2}+\mathbf{b}_{2},\label{ff}
\end{equation}
where $W_{1}, W_{2} \in \mathbb{R}^{d\times d}$ and $\mathbf{b}_{1},\mathbf{b}_{1}\in \mathbb{R}^{d}$ are learnable matrices.

\paragraph{Attention Mechanism.}
\label{Preliminary-Self-Attention} The term ``$\operatorname{Attention}$'' represents the Attention mechanism.
 
With the matrices $ X^{(l-1)} \in \mathbb{R}^{N\times d}$ and $C^{(l-1)}  \in \mathbb{R}^{M\times d}$ as input, the classical attention mechanism can be formulated as: 

\begin{equation}
\begin{aligned}
    & \operatorname{Attention}(X^{(l-1)} ,C^{(l-1)}) \\
    = &\operatorname{softmax}\left( \frac{ X^{(l-1)} W_Q (C^{(l-1)} W_K)^{\intercal}}{\sqrt{d}}\right)C W_V,
\end{aligned}
\label{attn}
\end{equation}

where $W_Q, W_K, W_V \in \mathbb{R}^{d\times d}$ are three learnable matrices. The attention function $\operatorname{Attention}(\cdot,\cdot)$ can integrate the information from $C^{(l-1)}$ to $X^{(l-1)}$ positionally.

The attention function has a computational complexity of $\mathcal{O}(N M)$ since each column of vector in $X^{(l-1)}$ interacts with every column of vector in $C^{(l-1)}$.

\paragraph{Self-attention} When $C^{(l-1)}=X^{(l-1)}$, Equation (\ref{attn}) becomes self-attention, which is with $\mathcal{O}(N^2)$ time and space complexity.

The constructive NCO model typically employs the self-attention~\citep{kwon2020pomo,drakulic2023bq,luo2023neural}. Their decoders need to perform $n$ steps for solution construction and each step takes $\mathcal{O}(n^2)$ runtime and space due to the execution of self-attention. Such high computational time and space complexity is the bottleneck for processing large-scale problems.

\section{Hyperparameters of Training Procedure}
\label{Hyperparameters of Training Procedure}
The warm-up model is trained on different scales of instances. On each scale, the dataset size, training epochs, local reconstruction batch size, and training batch size may be different. Here we provide the specific hyperparameters as shown in Table \ref{tab: training hyperparameters}. 

In each episode, we save the most improved model as the next training phase's model.

For each type of problem, the table lists the following hyperparameters:
\begin{itemize}

    \item \textbf{Dataset size.} It refers to the number of instances in the dataset, which varies depending on the problem scale. For example, the TSP100K dataset contains 50 instances, while both the TSP1K and CVRP1K datasets consist of 10,000 instances each. When dealing with larger scales, such as 100K, fewer instances are required for training because complete solutions at this scale can generate a wide variety of partial solutions, which are sufficient to contribute to effective training.
    
    \item \textbf{Episode.} It refers to one cycle of self-improved training. One episode involves a number of training epochs and a number of local reconstruction epochs. The values vary since the model's performance converges at different difficulties on different scales.
    
    \item \textbf{Training epoch.} An epoch refers to the process where the entire training dataset passes through the neural network once. In other words, when all training instances have been used for training the model once, one epoch is completed. It's consistent across different dataset sizes with a value of 20 for CVRP and TSP.
    
    \item \textbf{Training batch size.} It refers to the number of training data used each time the model parameters are updated. In one training epoch, the dataset is divided into several smaller batches, each containing a certain number of instances. Model parameter updates occur after each batch is processed by the network. During training, larger-scale datasets like TSP100K and CVRP100K use a smaller batch size due to device memory constraints, while smaller-scale datasets like TSP1K and CVRP1K use a larger batch size.
      
    \item \textbf{Local reconstruction epoch.} An epoch refers to the process where all training instances are reconstructed once. It's consistent across different dataset sizes with a value of 100 for CVRP and TSP.

    \item \textbf{Local reconstruction batch size.} In one local reconstruction epoch, the dataset is divided into several smaller batches, each containing a certain number of instances. Larger-scale datasets like TSP100K and CVRP100K use a smaller batch size due to device memory constraints, while smaller-scale datasets like TSP1K and CVRP1K use a larger batch size.
    
    \item \textbf{Learning rate.} The initial learning rate for training the model. It is set to 1e-4 across all problems.
    
    \item \textbf{Decay per training epoch.} The rate at which the learning rate decays with each epoch. The value is consistent at 0.97.
    
    \item \textbf{Partial solution’s maximum length.} This represents the maximum length for partial solutions considered during training and local reconstruction, which is set to $l_{max}=1000$ for training and inference efficiency.
\end{itemize}

\begin{table*}[h]
\centering
\caption{Training hyperparameters setting.}
\resizebox{0.6\textwidth}{!}{
\begin{tabular}{l|cccccc}
\toprule[0.5mm]
& TSP1K  & TSP5K   & TSP10K  & TSP50K  & TSP100K   \\ 
\hline
dataset size           & 10000  & 200  & 200   &  100   & 100  \\
Episode                   & 3      & 2    & 2     &  2    & 6 \\
\hline
Training batch size       & 256    & 32   & 32    &  32    & 16  \\
Training epoch           & 20     & 20   & 20    &  20    & 20 \\
\hline
Local reconstruction epoch   & 1   & 1 & 1    &  1 & 1  \\
Local reconstruction batch size & 1024   & 32   & 32    &  16    & 8  \\
local reconstruction iteration   & 100    & 100  & 100   &  100    & 100 \\
\hline
Learning rate            & 1e-4  & 1e-4 &1e-4  &  1e-4  & 1e-4   \\
decay per training epoch            & 0.97   & 0.97  & 0.97   &  0.97 
&  0.97    \\
\hline
Partial solution’s maximum length & 1000  & 1000 & 1000   &  1000  & 1000  \\
\hline
\hline
 & CVRP1K  & CVRP5K   & CVRP10K  & CVRP50K  & CVRP100K   \\ 
\hline
dataset size            & 10000  &  200   & 200   &  100   & 100  \\
Episode                & 11      & 9    &  15    &   2   & 3 \\
\hline
Training batch size      & 512    & 32   & 32    &  32    & 16  \\
Training epoch         & 20     & 20   & 20    &  20    & 20 \\
\hline
Local reconstruction epoch   & 1   & 1 & 1    &  1 & 1  \\
Local reconstruction batch size   & 1024   & 32   & 32    &  16    & 8  \\
local reconstruction iteration   & 100    & 100  & 100   &  100    & 100 \\
\hline
Learning rate            & 1e-4  & 1e-4 &1e-4  &  1e-4  & 1e-4   \\
decay per training epoch       & 0.97   & 0.97  & 0.97   &  0.97  &  0.97    \\
\hline
Partial solution’s maximum length & 1000  & 1000 & 1000   &  1000  & 1000  \\
\bottomrule[0.5mm]     
\end{tabular}
}
\label{tab: training hyperparameters}
\end{table*}

Note that the training time for these models can range from around one to ten days, depending on factors such as dataset size, the scale of the problem instance, and the episode, to ensure model performance convergence. However, RL-based methods, like POMO~\citep{kwon2020pomo}, typically require more extended training periods, sometimes up to two weeks. In contrast, our method not only demands less time but also demonstrates significantly better scalability, thereby highlighting its effectiveness and efficiency.

\section{Implementation Details}

\subsection{TSP}

\subsection{Problem setup}

Solving a TSP instance with $n$ nodes entails discovering the shortest loop that visits every node once and returns to the initial node. We follow the method in~\citet{kool2018attention} to create Euclidean TSP instances by sampling $n$ nodes' coordinates uniformly from a unit square $[0,1]^2$.

\subsection{Implementation}

\subsubsection{Encoder}

In a TSP instance $\mathbf{s}$, the node features $(\mathbf{s}_1,\ldots,\mathbf{s}_n)$ represent the 2D coordinates of the $n$ nodes in the graph. Given the node features $(\mathbf{s}_1,\ldots,\mathbf{s}_n)$, the encoder of our proposed model produces each node's embedding $\mathbf{h}_i, i= \{1,\dots,n\}$ by a linear projection:
\begin{eqnarray}  
\mathbf{h}_i = W^{(0)}\mathbf{s}_i  + b^{(0)},
\end{eqnarray}
where $W^{(0)} \in \mathbb{R}^{2 \times d}$ and $  b^{(0)} \in \mathbb{R}^{d}$ are learnable matrices.

\subsubsection{Decoder}
Assuming that at the construction step $t$, the number of starting nodes, destination nodes, and available nodes is $m$,
The output of the $L$-th linear attention module is 

\begin{equation}
    \widetilde{H}^{(L)}=\{\widetilde{\mathbf{h}}_i^{(L)}, i=1,\ldots,m\},
\end{equation}

which represents the relation of the whole graph nodes. Then, a linear projection and softmax function are applied to it, producing the selected probability of each available node. The starting and destination nodes are masked before the softmax calculation.
\begin{equation}
\begin{aligned}
        u_i &= \begin{cases}
           W_O\widetilde{\mathbf{h}}_{i}^{(L)}, &\text{$i\neq1$ or $2$}\\
           -\infty, &\text{otherwise}
           \end{cases},\\
        \mathbf{p}^t &= \operatorname{softmax}(\mathbf{u}), \ \mathbf{u}=\{u_i, i=1,\ldots,m\}.
\end{aligned}
\end{equation}
Where $W_O \in \mathbb{R}^{1 \times d}$ is a learnable matrix. The most suitable node $x_t$ is selected based on $\mathbf{p}^t = \{p_i, i=1,\ldots,m \}$.  Each $p_i,i\neq 1 \ \text{or} \ 2$ represents an available node's probability of being selected. Finally, a complete solution $\mathbf{x}=(x_{1},\ldots,x_{n})^\intercal$ is constructed by calling the decoder $n$ times.

\subsection{CVRP}

\subsection{Problem setup}

A CVRP instance comprises one depot node and $n$ customer nodes, where each customer node $i$ has a demand $\delta_i$ to fulfill. Our goal is to find a set of sub-tours that begin and end at the depot, ensuring the sum of demands in each sub-tour adheres to the vehicle's capacity constraint $D$. The objective is to minimize the total distance across these sub-tours while maintaining the capacity constraint $D$. Our CVRP instances generated similarly to~\citep{kool2018attention}, feature customer and depot node coordinate uniformly sampled from a unit square $[0,1]^2$. Demands $\delta_i$ are uniformly sampled from ${1, \dots, 9}$. Vehicle capacities $D$ are set to 250, 500, 1000, and 2000 for CVRP1K, CVRP5K, CVRP10K, and CVRP50K/100K respectively.

In line with~\citet{luo2023neural,drakulic2023bq,kool2022deep}, we establish a feasible CVRP solution formation. Instead of isolating a depot visit as a distinct step, we employ binary variables to signify whether a customer node is accessed via the depot or another customer node. In a feasible solution, a node is assigned $1$ if accessed through the depot and $0$ if accessed through another customer node. For example, a viable CVRP solution $\{0,1,2,3,0, 4, 5, 0, 6, 7, 0, 8, 9, 10\}$ with $0$ representing the depot can be represented as shown below:
\begin{equation}
\begin{bmatrix}
\label{CVRP solution notation}
1 & 2 & 3 & 4 & 5 & 6 & 7 & 8 & 9 & 10\\
1 & 0 & 0 & 1 & 0 & 1 & 0 & 1 & 0 & 0
\end{bmatrix}
\end{equation}
where the first row displays the visited node sequence, while the second row signifies if each node is accessed through the depot or another customer node.

This notation aims to maintain solution consistency. In CVRP cases, solutions with equal customer node counts might have differing sub-tour quantities, causing potential misalignment. This notation prevents such problems.

\subsection{Implementation}
\subsubsection{Encoder}
In CVRP, the node feature $\mathbf{s}_i$ is a 3D vector, combining 2D coordinates and node $i$'s demand. The depot's demand is set as 0. We normalize the vehicle capacity $D$ to $\hat{D}=1$ and the demand $\delta_i$ to $\hat{\delta}_i = \frac{\delta_i}{D}$ for simplicity~\citep{kool2018attention}.

Given the node features $(\mathbf{s}_1,\ldots,\mathbf{s}_n)$, the  encoder of our proposed model produces each node's embedding $\mathbf{h}_i, i=\{1,\dots,n\}$ by a linear projection:
\begin{eqnarray}  
\mathbf{h}_i = W^{(0)}\mathbf{s}_i  + \mathbf{b}^{(0)},
\end{eqnarray}
where $W^{(0)} \in \mathbb{R}^{3 \times d}$ and $  \mathbf{b}^{(0)} \in \mathbb{R}^{d}$ are learnable matrices.

\subsubsection{Decoder}

In the decoder, we add the dynamically changing remaining capacity to the starting and destination node embeddings, akin to~\citet{kwon2020pomo}. Denoted the remaining capacity as $d_r \in \mathbb{R}^{1}$, recall that the starting and destination node embeddings are $\mathbf{h}_{x_1}$ and $\mathbf{h}_{x_{t-1}}$, respectively, the information of remaining capacity is fused to the starting node and destination nodes' embeddings by:
\begin{eqnarray}  
\mathbf{h}^{\prime}_{x_1}&=& W_1[\mathbf{h}_{x_1},d_r]  + \mathbf{b}_1 \nonumber \\
\mathbf{h}^{\prime}_{x_{t-1}}&=&W_2[\mathbf{h}_{x_{t-1}},d_r ]  +\mathbf{b}_2,
\end{eqnarray}
Where $W_1,W_2 \in \mathbb{R}^{(d+1)\times d}$ and $\mathbf{b}_1, \mathbf{b}_2 \in \mathbb{R}^{d}$.
Then the initial embeddings of the representative points $R^{(0)}$ are calculated as below:
\begin{eqnarray}  
R^{(0)}&=&[\mathbf{h}^{\prime}_{x_1}, {\mathbf{h}^{\prime}_{x_{t-1}}}],
\end{eqnarray}

Similar to the case of TSP, the output of our model's $L$-th linear attention module is  $R^{(L)}$ and $\widetilde{H}^{(L)}$.  

Where \begin{equation}
    \widetilde{H}^{(L)}=\{\widetilde{\mathbf{h}}_i^{(L)}, i=1,\ldots,a+2\},
\end{equation}

which consists of the relation of the whole graph nodes. Then, a linear projection and softmax function are applied to it, producing the selected probability of each available node. The starting and destination nodes are masked before the softmax calculation.

\begin{equation}
\begin{aligned}
        \mathbf{u}_i &= \begin{cases}
           W_O\widetilde{\mathbf{h}}_{i}^{(L)}, &\text{$i\neq1$ or $2$}\\
           -\infty, &\text{otherwise}
           \end{cases},
\end{aligned}
\end{equation}
Where $W_O \in \mathbb{R}^{2 \times d}$ is a learnable matrix.

Each $\mathbf{u}_i \in \mathbb{R}^{2}$ corresponds to two actions for node $i$: accessed via the depot or another customer node. It corresponds to the notation in Equation (\ref{CVRP solution notation}).

Then $U=\{\mathbf{u}_i, i=1,\ldots,m\} \in \mathbb{R}^{m \times 2}$ is flattened to $\mathbf{u}^\prime \in \mathbb{R}^{2m}$.

Sequentially, a softmax function is applied to $\mathbf{u}^\prime$ to produce the selected probability of each action related to the corresponding available node:  

\begin{eqnarray}  
\mathbf{p}^t = \operatorname{softmax}(\mathbf{u}^\prime)
\end{eqnarray}
 $p_i$ and $p_{m+i}$, $i\neq 1 \ \text{or} \ 2$ indicate the probability of node $i$ being accessed via the depot or the starting node, respectively. 
 
 The most suitable node $x_t$ and the associated action (accessed via the depot or the starting node) is decided based on $\mathbf{p}^t$.  Finally, a complete solution $\mathbf{x}=(x_{1},\ldots,x_{n})^\intercal$ is constructed by calling the decoder $n$ times.

\end{document}